\documentclass[10pt,twocolumn]{article} 
\usepackage{simpleConference}
\usepackage{booktabs}       % professional-quality tables
\usepackage{multirow}
\usepackage{amsfonts}       % blackboard math symbols
\usepackage{nicefrac}       % compact symbols for 1/2, etc.
\usepackage{microtype}      % microtypography
\usepackage{xcolor}         % colors
\usepackage{comment}        % Comment
\usepackage{verbatim}       % Verbatim
\usepackage{algorithm}
\usepackage[noend]{algpseudocode}
\algrenewcommand\algorithmicdo{}
\algrenewcommand\algorithmicthen{}
\usepackage{makecell}
\usepackage{siunitx}
\usepackage{url}
\usepackage{amsmath}
\usepackage{bm, bbm}
\usepackage{amsthm}
\usepackage{mathtools}

\usepackage{enumitem}
\setlist{itemsep=0.2ex, topsep=1ex, partopsep=0ex, parsep=0.5ex, leftmargin=2.3ex}

\def\R{\mathbb{R}}

\def\indep{\perp \!\!\! \perp}

\def\GAM{GA$^2$M}

\begin{document}

\title{Interpretable Prediction and Feature Selection \\
for Survival Analysis}

\author{Mike Van Ness \\
Stanford University \\
Stanford, CA USA \\
mvanness@stanford.edu
\and
Madeleine Udell \\
Stanford University \\
Stanford, CA USA \\
udell@stanford.edu \\
}
\maketitle

\begin{abstract}
  Survival analysis is widely used as a technique to model time-to-event data when some data is censored,
  particularly in healthcare for predicting future patient risk.
  In such settings, survival models must be both accurate and interpretable so that users (such as doctors) can trust the model and understand model predictions. 
  While most literature focuses on discrimination, interpretability is equally as important.
  A successful interpretable model should be able to describe how changing each feature impacts the outcome, and should only use a small number of features.
  In this paper, we present DyS (pronounced ``dice''), a new survival analysis model that achieves both strong discrimination and interpretability.
  DyS is a feature-sparse Generalized Additive Model, combining feature selection and interpretable prediction into one model.
  While DyS works well for all survival analysis problems, it is particularly useful for large (in $n$ and $p$) survival datasets such as those commonly found in observational healthcare studies.
  Empirical studies show that DyS competes with other state-of-the-art machine learning models for survival analysis, while being highly interpretable.
\end{abstract}

\section{Introduction}

% Story:
% \begin{itemize}
%     \item Survival analysis is an important problem, notably for healthcare applications.
%     \item Many ML methods proposed, but Cox model still most widely accepted because of easy interpretability.
%     \item XAI popular field in normal ML, but still studied for survival analysis, which has its own set of challenges.
%     \item Some existing approaches for XAI in survival analysis, but all have limitations. Most notably, none can combine feature selection and prediction, which is critical for large data settings.
%     \item Build a new model, called DyS, that performs feature selection and interpretable prediction for survival analysis data.
% \end{itemize}

% Contributions:
% \begin{itemize}
%     \item New accurate yet interpretable survival model equipped for big data.
%     \item First model to offer feature selection or feature-sparse predictions without assuming PH on survival data.
%     \item First interpretable survival model to scale to large data while having comparable discrimination to non-interpretable models (surv-stacking requires pre-feature selection, PH models are worse, and PseudoNAM performance is bad).
%     \item (similar to above) first to show that interpretable models can achieve competitive performance on survival analysis data, which is already known in classification but survival analysis is harder.
% \end{itemize}

% Contribution key words
% \begin{itemize}
%     \item Interpretable but performant.
%     \item survival analysis.
%     \item Scalable.
%     \item Feature selection.
% \end{itemize}

Predicting the time until an event occurs is a classic and important problem in many domains, including healthcare, customer churn, and machine failure. 
One solution for such time-to-event prediction is using regression techniques that support a strictly positive response.
However, many such time-to-event problems face the additional challenge of \textit{censoring}, where the event is never reached for a portion of samples.
When the censoring rate is low, removing all censored samples and using regression may be reasonable.
Often, though, the censoring rate is quite high (e.g. in healthcare), in which case removing censored samples loses valuable signal. 
Thus, an alternative approach is necessary in order to elegantly handle time-to-event data with censoring.

Survival analysis \cite{jenkins2005survival} is the standard and widely-adopted alternative approach for time-to-event data with censoring.
Survival analysis models aim to estimate the conditional distribution of time-to-event given a collection of features, using loss functions that allow learning from both censored and uncensored patients.
Statistical approaches for survival analysis, including the Cox proportional hazards model \cite{cox1972regression} and the accelerated failure time model \cite{cox1984analysis}, are traditional models in the field.
More recently, a plethora of machine learning \cite{wang2019machine} and deep learning \cite{wiegrebe2023deep} models have been proposed for survival analysis, most of which are shown to outperform traditional statistical models in terms of discrimination. 
Nonetheless, many fields have yet to fully adopt machine learning techniques for survival analysis, instead favoring classical statistical approaches due to their simplicity and inherent interpretability.

The field of Explainable AI (XAI) aims to bridge this interpretability gap between simple statistical approaches and more powerful machine learning approaches.
% Outside of survival analysis, there has been an increasing effort to develop explainable AI (XAI) methods for traditional machine learning tasks.
% These XAI approaches aim to bridge the gap in interpretability between traditional statistical approaches and more powerful machine learning approaches. 
Many XAI methods are designed to explain the behavior of black-box machine learning methods as a post-processing step \cite{lundberg2017unified, ribeiro2016should}; however, these methods are prone to approximation bias \cite{kumar2020problems, van2022tractability, alvarez2018robustness, rahnama2019study}.
More promising are inherently interpretable machine learning models, or glass-box models, that offer explanations without requiring post-hoc approximation.
These glass-box methods have surprisingly been shown to be competitive with black-box models for classification and regression on tabular datasets \cite{nori2019interpretml}.
Nonetheless, much less research has focused on XAI, and specifically glass-box modeling, for survival analysis. 
Some previous methods have been proposed \cite{van2023interpretable, xu2023coxnam, rahman2021pseudonam, peroni2022extending}, but no existing glass-box survival model is able to both compete with black-box survival models as well as scale to large and high-dimensional survival datasets.

To fill this gap, we propose a novel glass-box machine learning model for survival analysis, called DyNAMic Survival, or simply DyS (pronounced ``dice'').
DyS, like many other glass-box machine learning models, is a Generalized Additive Model (GAM) with additional shape functions for feature interactions.
Unlike previous glass-box survival models, though, DyS is trained using a ranked probability score (RPS) loss function that directly optimizes the survival predictions, leading to better discrimination \cite{kamran2021estimating, avati2020countdown}.
Additionally, DyS can perform feature selection during the model fitting process, both on the main effects and on the interaction terms, which previous glass-box survival models are incapable of doing.
% Such feature selection is crucial for high dimensional, as 1) feature correlation, which is reduced by feature selection, dampens the explained impact of features \cite{liu2021controlburn} 2) high dimensional data presents memory and computational challenges without feature selection, especially when modeling feature interactions 3) feature selection is sometimes necessitates by problem context. 
To summarize our main contributions:
\begin{itemize}
    \item We present a new survival analysis model, DyS, which achieves competitive discriminative performance while being a glass-box model. DyS can generate feature importances as well as feature impact plots at specific evaluation times (see Figure \ref{fig:interp_plots}) without requiring post-hoc approximation.
    \item We show how DyS can also be used for nonlinear feature selection on survival data. Such feature selection can be done as a preprocessing step, or can be integrated directly into the prediction model to generate feature-sparse intepretable predictions.
    \item We introduce a two-stage fitting approach which, when combined with feature-sparsity, allows DyS to scale to large survival analysis problems, where other approaches are either too slow and/or require separate feature selection as a preprocessing step.
\end{itemize}

% \begin{itemize}
%     \item New accurate yet interpretable survival model equipped for big data.
%     \item First model to offer feature selection or feature-sparse predictions without assuming PH on survival data.
%     \item First interpretable survival model to scale to large data while having comparable discrimination to non-interpretable models (surv-stacking requires pre-feature selection, PH models are worse, and PseudoNAM performance is bad).
%     \item (similar to above) first to show that interpretable models can achieve competitive performance on survival analysis data, which is already known in classification but survival analysis is harder.
% \end{itemize}

% Contribution key words
% \begin{itemize}
%     \item Interpretable but performant.
%     \item survival analysis.
%     \item Scalable.
%     \item Feature selection.
% \end{itemize}

\subsection{Example Usage}

Suppose a medical researcher wishes to study the risk factors associated with heart failure.
There are two common approaches to such a problem:
1) the researcher uses their medical expertise to select a small set of features they believe to be important risk factors for heart failure, and then builds a Cox proportional hazards model using these selected features. 
The researcher can use the regression coefficients of the Cox model, and their associated p-values, to determine which risk factors are most associated with heart failure.
2) the researcher gathers data for all possible features, and uses a large black-box survival model to predict heart failure risk. 
The model achieves state-of-the-art discrimination, 
but requires a post-hoc approximation method like SHAP to interpret the model's predictions.

\begin{figure*}
    \centering
    \begin{minipage}{.5\textwidth}
        \centering
        \includegraphics[width=\textwidth]{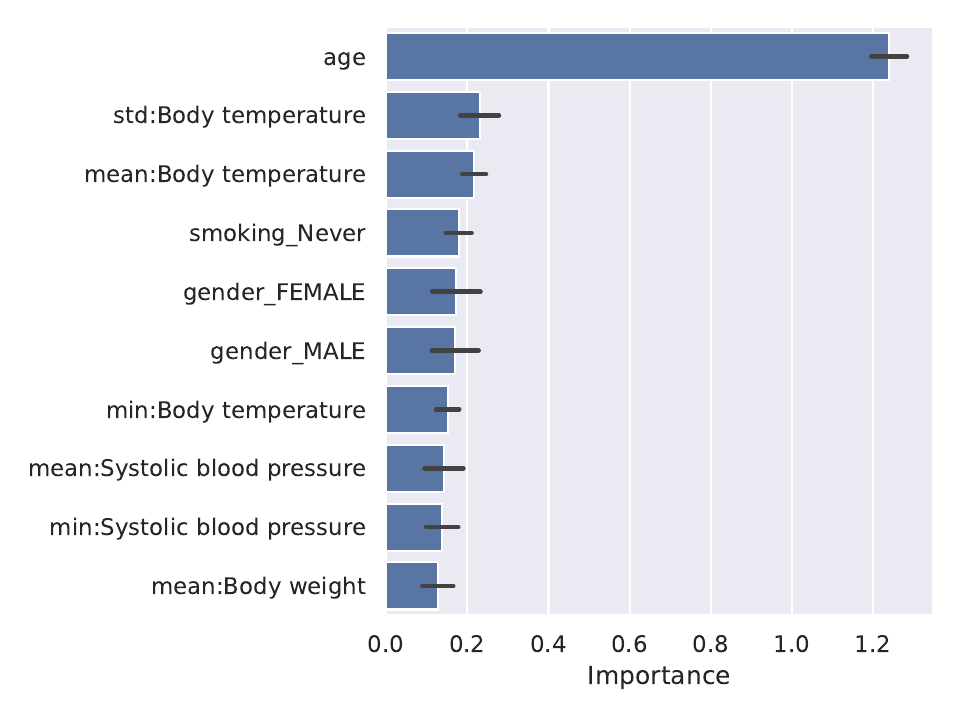}
    \end{minipage}%
    \begin{minipage}{.5\textwidth}
        \centering
        \includegraphics[width=\textwidth]{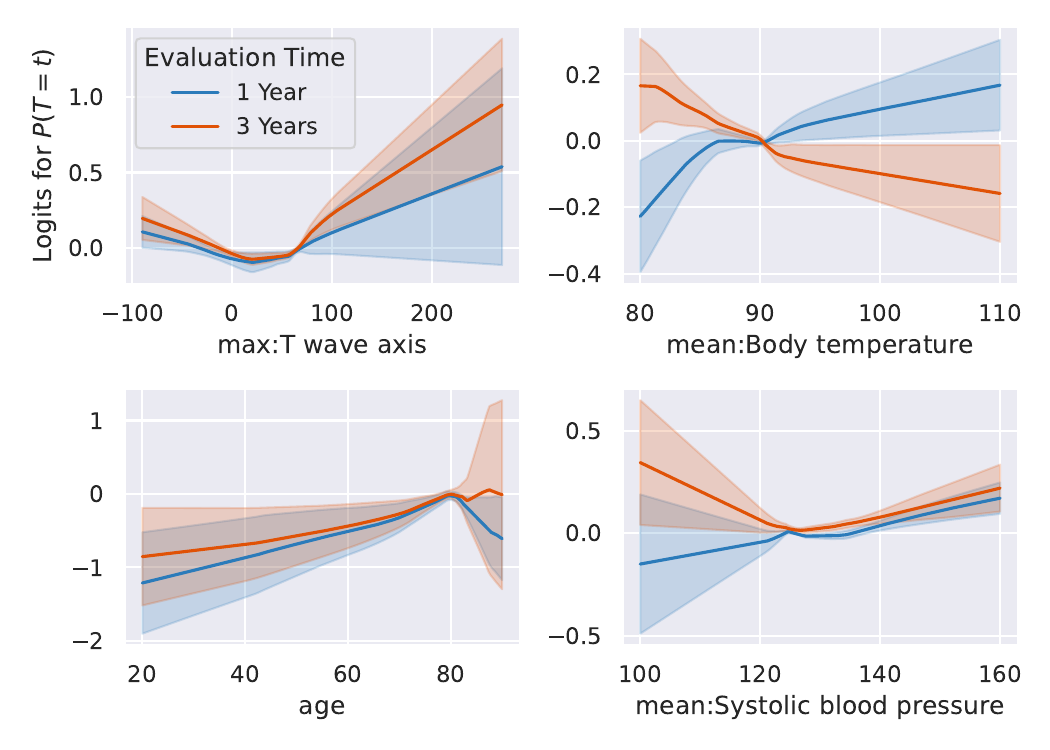}
    \end{minipage}
    \begin{minipage}{.5\textwidth}
        \centering
        \includegraphics[width=\textwidth]{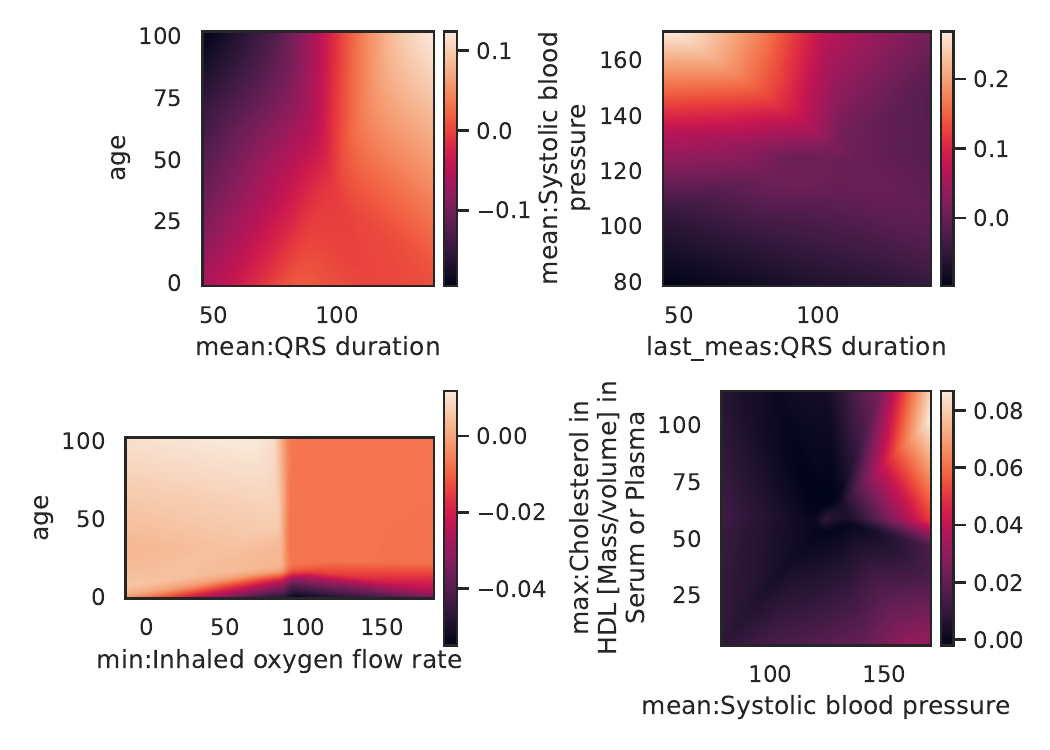}
    \end{minipage}%
    \begin{minipage}{.5\textwidth}
        \centering
        \includegraphics[width=\textwidth]{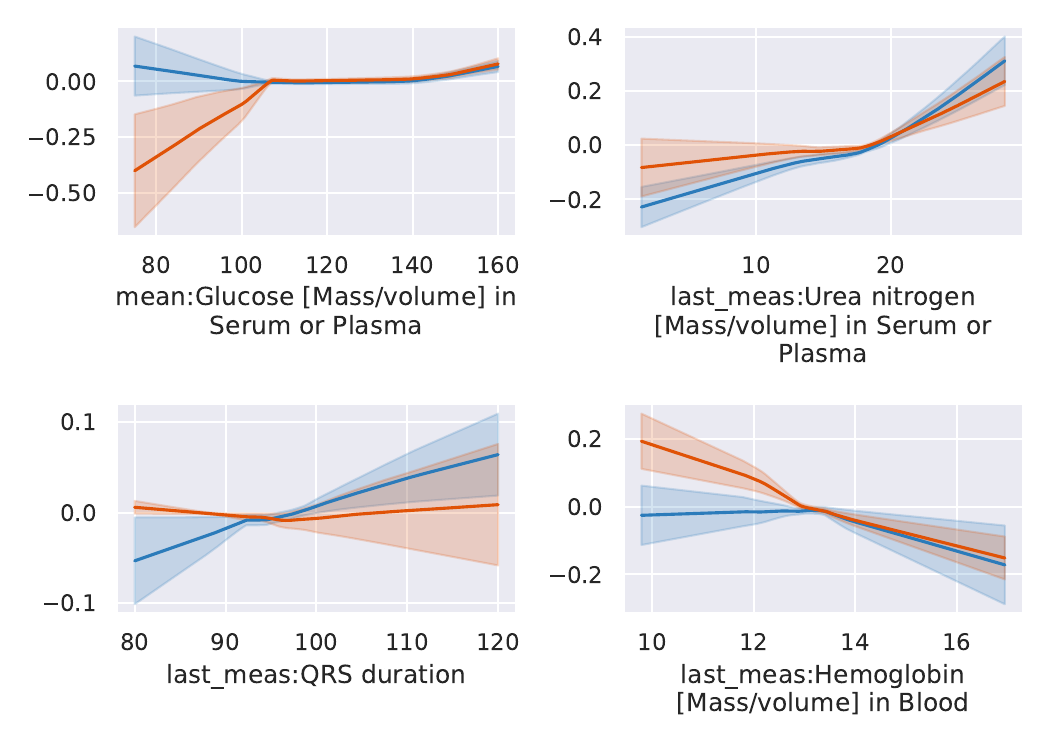}
    \end{minipage}
    \caption{Interpretable plots generated by DyS trained on heart failure data, across 10 trials. (Top left) feature importances averaged across evaluation times. (Right) feature impact plots for individual features at two evaluation times: 1 year and 3 years. (Bottom left) feature impact plots for interactions at 1 year. These plots fully describe the behavior of the fitted DyS model without any extra processing due to DyS's glass-box structure. }
    \label{fig:interp_plots}
\end{figure*}

Instead, the researcher can use DyS to achieve the benefits of both approaches.
To illustrate, we fit DyS on a real heart failure dataset as described in Section \ref{sec:hf}
to produce the interpretability plots shown in Figure \ref{fig:interp_plots}.
The feature importance plot (left) directly shows the researcher which features are most important for heart failure prediction, 
and the feature impact plots (right) show how changes in each feature change the risk score at specific times in the future.
From these plots, the researcher notices a surprising pattern in some feature, and uses this observation to guide further research.

\section{Background}

\subsection{Survival Analysis}
\label{sec:survival_analysis}

Survival analysis is a standard approach for modeling time-to-event data with censoring.
Survival analysis is most often used for modeling healthcare data, 
and has many other applications such as predicting equipment failure \cite{papathanasiou2023machine}, economics \cite{danacica2010using}, and customer churn \cite{lariviere2004investigating}. 

Mathematically, survival analysis consists of data of the form $(X, T, \delta)$.
Here $X \in \R^p$ represents a vector of $p$ features, 
$T \in [0, \infty)$ represents the event time, 
and $\delta \in \{0, 1\}$ represents the censor indicator. 
When $\delta = 1$, the sample is considered \textit{uncensored}, 
and the time $T$ represents the time which the event of interest occurs. 
When $\delta = 0$, the sample is instead called \textit{censored}, 
in which case the value of $T$ is unknown and is replaced with the last available observation time.
There are a variety of reasons in which samples could be censored;
in healthcare setings, a patient could switch hospital networks, 
die of unrelated causes, or not be diagnosed with heart failure before the end of the study.
% As in traditional supervised machine learning, we observe a training dataset of $n$ samples $\{X^{(i)}, T^{(i)}, \delta^{(i)})\}$, from which we would like to use to train a 

The goal of survival analysis is to estimate the conditional distribution of $T$ given $X$. 
Often, the quantity directly of interest is the \textit{survival function}: 
\begin{equation}
S(t \mid X) = P(T > t \mid X). 
\end{equation}
Some models instead predict the \textit{hazard function} $\lambda(t \mid X)$, which is the conditional density at $T = t$ conditioned on surviving to at least time $t$:
\begin{equation}
    \lambda(t \mid X) = p(t \mid T \geq t, X) = \frac{p(t)}{S(t \mid X)}.
\end{equation}
For example, the most traditional model in survival analysis is the Cox proportional hazards models \cite{cox1972regression}, which fits a linear function to the log hazard function:
\begin{equation}
    \lambda(t \mid X) = \lambda_0(t) \exp(X^T \beta).
\end{equation}
Here $\lambda_0(t)$ is called the baseline hazard function, while $\exp(X^T \beta)$ gives the offset to the baseline hazard for each sample. 
This factorized approach results in predictions with \emph{proportional hazards}: 
for samples $X, X'$, their hazards $\lambda(t \mid X)$ and $\lambda(t \mid X')$ are proportional and independent of $t$:
\begin{equation}
\label{eq:ph_assumption}
    \frac{\lambda(t \mid X)}{\lambda(t \mid X')} = \frac{\exp(X^T \beta)}{\exp(X'^T \beta)} \indep t.
\end{equation}
Several machine learning models have been proposed to fit proportional hazards models by replacing $f(X) = X^T \beta$ with some nonlinear $f$ \cite{katzman2018deepsurv, hastie1986generalized, ridgeway1999state}.
Further, several papers have proposed methods to overcome the proportional hazards assumption, particularly using deep learning \cite{lee2018deephit, rahman2021deeppseudo, bennis2020estimation}. 
Most of these methods, though, are not interpretable; thus,
an important goal of this paper is to develop an interpretable machine learning model for survival analysis that does not assume proportional hazards.

\subsection{Interpretable Machine Learning}

The field of Explainable AI (XAI) aims to build methods that explain the behavior of machine learning models.
These explanations can take many forms.
In this paper, we aim to deliver the following kinds of model explanations:
\begin{itemize}
    \item \textbf{Feature importances}: feature importances demonstrate which features most impact the model's performance. Typically, each feature is assigned a single importance score, with a higher score indicating a more important feature.
    \item \textbf{Feature impact plots}: feature impact plots demonstrate how changing a particular feature impacts the model's predictions. Feature impact plots can be used to assess the relationship between individual features and the response, which can either confirm existing understanding or lead to new discoveries about feature/response relationships.
\end{itemize}

XAI comprises two fundamentally different kinds of interpretability: post-hoc and glass-box.
Celebrated post-hoc approximation methods have been developed to explain black box machine learning models \cite{dwivedi2023explainable}.
In contrast, this paper instead focuses on glass-box machine learning, aiming to build models that provide natural explanations (both importance and impact plots) without the need for post-hoc approximations.
While glass-box machine learning has a long history \cite{longo2020explainable}, 
only recently have 
they been shown to deliver performance comparable to modern black-box machine learning models \cite{nori2019interpretml}.

% Many write something about differences between different ways of getting feature importances and feature impact plots...

% Feature impact plots are a kind of partial dependence plot. They differ in that the feature impact plot completely summarized the behavior of the model, whereas a partial dependence plot only approximates it (sometimes, locally).

Most modern glass-box machine learning models are based on Generalized Additive Models (GAMs) \cite{hastie1986generalized}. For a data vector $X$ with $p$ features, a GAM $f$ is formulated as
\begin{equation}
\label{eq:gam}
    f(X) = \sum_{j=1}^p f_j(X_j).
\end{equation}
The individual feature functions $f_j$ are sometimes referred to as \textit{shape functions}.
The GAM $f$ can be optimized with respect to any loss function, e.g. the cross entropy loss for classification.
If $f_j(X_j) = X_j \beta_j$, the GAM is a generalized linear model. 
However, allowing nonlinear shape functions $f_j$ typically results in more accurate models.
More recently, many methods have explicitly added interaction terms to GAMs \cite{lou2013accurate}, 
resulting in so-called GA$^2$Ms:
\begin{equation}
\label{eq:ga2m}
    f(X) = \sum_{j=1}^p f_j(X_j) + \sum_{j \neq \ell} f_{j, \ell}(X_j, X_\ell).
\end{equation}
In these models, functions of a single feature are called \textit{main effects}, 
while functions of pairs of features are called \textit{interactions}. 
Different architectures can be used to fit each of the shape functions.
While splines have historically been the most popular \cite{hastie1986generalized}, 
recent works have used decision trees \cite{lou2012intelligible, lou2013accurate} 
and neural networks \cite{agarwal2021neural, chang2021node, ibrahim2023grand, chang2022data}.

\subsubsection{Feature Selection}
\label{sec:feature_selection}

An unappreciated yet important part of interpretability is feature selection.
When the number of features in a dataset is large, the data can generally be approximated well by a low-rank matrix \cite{udell2019big}; as a consequence, many features are likely to be highly correlated.
Feature correlation can negatively impact model interpretability, as sets of correlated features can dampen each other's feature importance and feature impact plots, making any individual feature in the set appear less important \cite{liu2021controlburn}.
% They can also result in unintuitive changes to feature impact plots. % citation???
Feature selection has additional benefits as well: 
models with fewer features are easier to explain to domain experts, 
require fewer model parameters, 
and are easier to generalize to new datasets.
In extreme cases when the number of features is larger than the number of samples (e.g. in genomics), 
feature selection may be necessary for an interpretable model to perform well.

\section{Methodology}

We now describe DyS, a novel glass-box machine learning model for survival analysis. 
DyS is designed to fill existing holes in the literature for glass-box survival analysis.
Specifically, DyS has the following properties:
\begin{itemize}
    \item \textbf{Interpretable.} DyS is a glass-box machine learning method, providing explanations inherently due to the model structure.
    \item \textbf{Performant.} DyS performs on par with state-of-the-art methods for survival analysis. 
    \item \textbf{Scalable.} DyS can handle large datasets: for example, a dataset with 2000 features and 500,000 samples can be fit in less than an hour with 1 GPU and 30 GB of RAM. 
    % compute time and memory usage, for large survival analysis (both in number of features and number of samples).
    \item \textbf{Feature-sparse.} Last, DyS can automatically select features during fitting, yielding improved interpretability for datasets with many features (see Section \ref{sec:feature_selection}) without requiring separate feature selection as preprocessing. 
    % Feature selection is important when the number of features is large or when features may be highly correlated. 
    Thus, DyS can be used for feature selection as well as feature-sparse prediction.
\end{itemize}

\begin{algorithm}
\caption{DyS Survival Predictions}
\begin{algorithmic}[1]
    \State \textbf{Input}: 
    \begin{itemize}[noitemsep, topsep=0pt]
        \item Evaluation times $t_1, \ldots, t_K$.
        \item Feature logits $f_j(X_j) \in \R^{K}$.
        \item Interaction logits $f_{j, \ell}(X_j, X_\ell) \in \R^K$.
    \end{itemize}

    \vspace{2pt}

    % \For{$ k = 1, \ldots, K$}
    \State $f(X) \gets \sum_j f_j(X_j) + \sum_{j, \ell} f_{j, \ell}(X_j, X_\ell)$.
    % \EndFor
    \State $\{\hat{P}(T = t_k \mid X)\}_{k=1}^K \gets \text{softmax} (f(X)) \in (0,1)^K$.
    \For{$ k = 1, \ldots, K$}
        \State $\hat{S}(t_k \mid X) \gets 1 - \sum_{t \leq t_k} \hat{P}(T = t \mid X)$.
    \EndFor
        
    \State \Return Survival predictions $\hat{S}(t_k \mid X),~ k=1, \ldots, K$.
    
\end{algorithmic}
\label{alg:prediction}
\end{algorithm}

\subsection{Model Architecture}

\begin{figure}
    \centering
    \includegraphics[width=\linewidth]{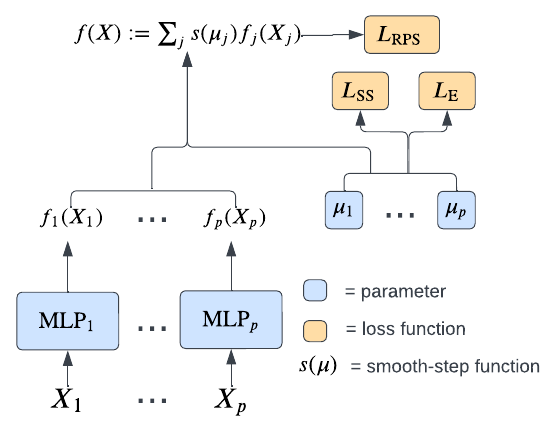}
    \caption{Summary of the architecture of DyS. For simplicity, the interaction effects are not shown. When feature sparsity is desired, the $\mu_j$ parameters are learned such that a subset of $s(\mu_j), j = 1, \ldots, p$ are equal to 0, preventing the corresponding features from influencing the predictions.  }
    \label{fig:model_arch}
\end{figure}

The model architecture for DyS is summarized in Figure \ref{fig:model_arch}.
Following previous glass-box models, DyS is a generalized additive model with interactions, or a GA$^2$M model.
We choose to parameterize each shape function in DyS using neural networks, specifically MLPs, making DyS a neural additive model (NAM) with interactions, or a NA$^2$M. 
% We elect not to use gradient boosted trees for the shape functions to allow for additionally flexibility in feature selection and regularization, as we describe in the proceeding sections.

DyS uses a discrete-time model, summarized in Algorithm \ref{alg:prediction}, to generate survival predictions.
We assume $T$ is discrete with finite support at $K$ distinct times: $T \in \{t_1, t_2, \ldots, t_K\}$. 
Under this model, DyS produces $K$-dimensional outputs $f(X_j), f(X_j, X_\ell) \in \R^K$, for each feature $X_j$ and interaction $(X_j, X_\ell)$. 
The global $K$-dimensional output $f(X)$ is obtained by summing the outputs from all main effects and interactions, as in Equation \ref{eq:ga2m}.
Next, a softmax layer computes probability mass estimates $\hat{P}(T = t_k \mid X)$ for $k = 1, \ldots, K$. 
The probability of survival to time $t_k$, $\hat{S}(t_k \mid X)$, is one minus the sum of the probability mass estimates at earlier times.

\subsubsection{Interpretation Plots}

DyS can be fully summarized by feature (and feature interaction) impact plots, one for each evaluation time; see Figure \ref{fig:interp_plots}.
Specifically, the feature impact plot for feature $j$ shows the logits of $P(T = t \mid X_j)$ at each evaluation time $t$.
Given a set of features $X$, the model's survival predictions $\hat{S}(t \mid X)$ can be obtained directly from these plots by following Algorithm \ref{alg:prediction}.
DyS's feature impact plots are critically different than partial dependence plots \cite{greenwell2017pdp}, as DyS's feature impact plots directly describe the model behavior using only the model structure.

DyS can also compute the importance of each feature $j$ at each time $t$ by averaging the absolute value of the logits of feature $j$ and time $t$ across all training samples.
If global feature importances are desired (i.e. independent of evaluation time), 
the importances at each evaluation time can further be averaged.

\subsection{Loss Function}
\label{sec:loss_fn}

% Recall that a GAM fits a model $f(X)$
% Motivate why we need something to map from a continuous score to a loss in order to hook up GAMs with survival analysis.

To train survival models, we must map the continuous prediction function $f(X)$ to a differentiable loss function.
Choosing an appropriate loss function for survival analysis is critical.
While traditional supervised machine learning problems often use a standard loss functions (MSE for regression, cross-entropy for classification), there is no standard loss function for survival analysis.

The most commonly used loss function in survival analysis is the Cox proportional hazards loss:
\begin{equation}
\label{eq:cox_loss}
    - \prod_{i : \delta^{(i)} = 1} \frac{\exp(f(X^{(i)}))}{\sum_{j : T^{(j)} \geq T^{(i)}} \exp(f(X^{(j)}))},
\end{equation}
where $(X^{(i)}, T^{(i)}, \delta^{(i)})$ is the $i^{\text{th}}$ sample in the training dataset.
This loss was originally derived as a negative partial likelihood function for the Cox model \cite{cox1972regression} (in the Cox model, $f(X) = X^T \beta$).
Intuitively, the loss works by maximizing the risk $\exp(f(X^{(i)}))$ for uncensored samples $i$ at their event-times $T^{(i)}$, relative to the other samples $j : T^{(j)} \geq T^{(i)}$ still at risk at time $T_i$. 
In practice, the log of Equation \ref{eq:cox_loss} is usually optimized for numerical stability.

There are two major problems with the Cox loss.
First, the Cox loss enforces proportional hazards, as illustrated in Equation \ref{eq:ph_assumption}. 
This property asserts that if the model predicts a higher risk for sample $i$ than sample $j$ at time $t$, 
it also predicts a higher risk for sample $i$ at all other times, which may be unreasonable.
% the model ranks the risk of a sample the same across times.
For example, consider mortality prediction:
a childhood cancer patient is at much higher risk of mortality than a typical 60 year old patient,
but if the child lives into adulthood, their risk may drop relative to the older patient.
We revisit this example empirically in Section \ref{sec:exp_synth}. 
Second, the Cox loss (or anything similar, e.g. the time-dependent Cox loss \cite{therneau2017using}) takes as input risk predictions rather than survival predictions. 
Thus, models that use the Cox loss must transform the model predictions in order to obtain survival predictions, which are usually the quantity of interest.
% This is particularly sub-optimal for interpretable models, where it is greatly beneficial to be able to interpret the model predictions directly related to the survival predictions.

For these reasons, we choose the Ranked Probability Score (RPS) loss function \cite{kamran2021estimating} instead of a Cox-based loss. 
First, DyS follows Algorithm \ref{alg:prediction} to generate a survival curve estimate $\hat{S}(t \mid X)$ from the prediction function $f(X)$.
Then the RPS loss is calculated as
\begin{equation}
\label{eq:rps_loss}
L_{\text{RPS}}(\hat{S}, T, \delta) = \sum_{t < T} (1 - \hat{S}(t \mid X))^2 + \delta \sum_{t \geq T} \hat{S}(t \mid X)^2
\end{equation}
for a finite set of evaluation times $t \in (0, \max(T))$. 
The RPS loss is a discrete-time version of the Continuous Ranked Probability Score \cite{gneiting2007probabilistic, avati2020countdown} originally proposed for time series forecasting.
The first term in $L_{\text{RPS}}$ maximizes survival for all samples before their respective event time, while the second term in $L_{\text{RPS}}$ minimizes survival for all uncensored samples after their event time.
Compared to Cox-style losses, the RPS loss directly optimizes survival predictions, the usual quantity of interest.

\subsection{Feature Sparsity}
\label{sec:feature_sparsity}

To obtain a feature-sparse model, inspired by \cite{ibrahim2023grand}, we introduce binary gates into our GA$^2$M model:
\begin{equation}
    f(X) = \sum_j f_i(X_j)z_j + \sum_{j,\ell} f_{j, \ell} (X_j, X_\ell) z_{j, \ell}.
\end{equation}
Each binary gate $z_j, z_{j, \ell} \in \{0, 1\}$ controls whether or not each feature/interaction is used by the final model. 
The binary gates are learned as parameters in the model alongside the shape functions $f_j, f_{j, \ell}$. 
Given the discrete nature of the binary gates, we follow \cite{ibrahim2023grand} and replace the binary gates with smooth-step functions $s(\mu_j), s(\mu_{j, \ell}) \in [0, 1]$ with real-valued parameters $\mu_j, \mu_{j, \ell} \in \R$ \cite{hazimeh2020tree}.
The smooth-step function is a continuous function with range $[0, 1]$, but unlike other such functions like the sigmoid function, the smooth-step function can actually reach 0, producing an exactly sparse model (see Appendix \ref{sec:smooth_step}).
The resulting GA$^2$M model is
\begin{equation}
    f(X) = \sum_j f_j(X_j)s(\mu_j) + \sum_{j,\ell} f_{j, \ell} (X_j, X_\ell) s(\mu_{j, \ell}).
\end{equation}
To induce feature sparsity, the smooth-step parameters $\mu_j, \mu_{j, \ell}$ are learned via the RPS loss $L_{RPS}$ along with the sparsity regularizer $ L_{\text{SS}}(\mu)$:
\begin{equation}
\label{eq:ss_loss}
    L_{\text{SS}}(\mu) = \lambda \left( \sum_j s(\mu_j) + \alpha \sum_{j, \ell} s(\mu_{j, \ell}) \right).
\end{equation}
Last, to control the rate at which the smooth-step functions converge to $\{0, 1\}$, we add the entropy regularizer $L_{\text{E}}(\mu)$:
\begin{align}
    L_{\text{E}}(\mu) &= \tau \left( \sum_j \Omega(s(\mu_j)) + \sum_{j,\ell} \Omega(s(\mu_{j, \ell}))\right), \\
    \Omega(x) &= - \Big( x \log(x) + (1 - x) \log(1 - x) \Big)
\end{align}
where $\Omega(x)$ explicitly encourages each $s(\mu_j), s(\mu_{j, \ell})$ to converge to 0 or 1. DyS is trained by minimizing $L_{RPS} + L_{SS} + L_E$ with respect to the trainable parameters in each shape function as well as the smooth-step parameters.

\subsubsection{Preset Feature Budget}

Occasionally, a user may want a model that uses a fixed number of features. 
This can be achieved by finding a hyperparameter $\lambda$ that yields the desired number of non-zero features.
One algorithmic way to select such a hyperparameter is \emph{bisection}, 
which uses a binary search algorithm to converge to a feasible hyperparameter value;
see Algorithm \ref{alg:bisection} in Appendix \ref{sec:bisection}.
We will use this algorithm later in Section \ref{sec:select_k} to compare feature selection methods.

\begin{algorithm}
\caption{DyS Two-Stage Fitting}
\begin{algorithmic}[1]
    \State \textbf{Input}: Survival dataset $D$.

    \State Fit main effects of DyS model $f$ on D using $L_{\text{RPS}} + L_{\text{SS}} + L_{\text{E}}$.

    \State Freeze existing parameters of $f$.

    \State Gather candidate interactions $(X_j, X_\ell)$ such that $s(\mu_j) > 0$ and $s(\mu_\ell) > 0$ in $f$.
    % Add new parameters to $f$ for candidate interactions.

    \State Fit parameters in $f$ for candidate interactions. 
    % adding predicted logits for main effects and interactions to generate survival probabilities.
        
    \State \Return Fitted DyS model $f$.
    
\end{algorithmic}
\label{alg:two-stage}
\end{algorithm}

\subsection{Two-Stage Fitting}
\label{sec:two-stage}

Training \GAM s is computationally and memory intensive when the number of features in a dataset is large.
For example, a dataset with $p = 1,000$ features contains $\sim 500,000$ possible interactions.
Previous literature has suggested an initial round of \textit{interaction screening} in order to reduce the number of interactions in the model: EBM \cite{lou2013accurate} uses an efficient plane cutting algorithm, while Grand-Slamin' \cite{ibrahim2023grand} fits shallow decision trees for all possible interactions to find the most promising interactions.
Unfortunately, we found even these efficient methods to be too inefficient when both $n$ and $p$ are large.
For example, on our heart failure dataset with training data of size $(n, p) = (537836, 2410)$ (see Section \ref{sec:hf}), the estimated run time for fitting a single decision tree on all possible interactions (as in \cite{ibrahim2023grand}) is over 100 hours.

Instead, DyS takes advantage of the feature sparsity in the model.
We first fit only the main effects of the model. 
The fitting procedure naturally chooses an active subset of the main effects by learning the smooth-step function parameters.
Then, we freeze the main effects and fit interaction effects only for interactions between two active main effects. 
This two-stage fitting approach, summarized in Algorithm \ref{alg:two-stage},
has both computational and interpretability benefits: 
fitting main effects first ensures that each main effect captures the entire available signal
and does not leak into the interaction shape function.
Hence, the shape function for each main effect represents the ``pure'' effect of each feature \cite{lengerich2020purifying}. % resulting in better interpretability.

\section{Related Work}

Interpretable prediction and feature selection both have long histories in machine learning for classification and regression. 
In the broad field of Explainable AI (XAI), many post-hoc methods for approximating feature importance and feature impact plots have been proposed, most notably SHAP \cite{lundberg2017unified} and LIME \cite{ribeiro2016should}. 
Other previous works have proposed naturally interpretable models, mostly using GAMs \cite{hastie1986generalized}.
Explainable Boosting Machines \cite{lou2013accurate, nori2019interpretml} fit GAMs with interactions for interpretable ML using gradient boosted decision trees, 
while neural additive models \cite{agarwal2021neural} and NODE-GAM \cite{chang2021node} use neural networks as shape functions.
For feature selection, many approaches utilize L1 penalization \cite{tibshirani1996regression} with either tree-based models \cite{liu2021controlburn, deng2012feature} or neural networks \cite{dinh2020consistent, zhang2019feature, wang2020feature}.
Much rarer are machine learning models capable of interpretable prediction and feature selection, with a few works proposed very recently \cite{ibrahim2023grand, xu2023sparse}. 

Separately, many machine learning models have been proposed for survival analysis, albeit most are not interpretable nor can perform feature selection.
Ensembles of decision trees have been adapted to survival analysis, including Random Survival Forests \cite{ishwaran2008random} and gradient boosting methods \cite{hothorn2006survival, chen2013gradient, bai2022gradient}.
Further, many deep learning approaches have been proposed for survival analysis.
These deep learning approaches largely fall into three categories: models which use some form of the Cox loss \cite{katzman2018deepsurv, zhong2022deep, nagpal2021deep}, models which fit a parametric distribution to the time-to-event variable \cite{bennis2020estimation, avati2020countdown, bennis2021dpwte}, and models which fit a discrete distribution to the time-to-event variable \cite{lee2018deephit, gensheimer2019scalable}.
Our paper is most similar to the third category, as we use a discrete-time approach in DyS.

Last, some previous work has explored interpretable prediction or feature selection for survival data.
Using GAMs for interpretable survival analysis was first described in \cite{hastie1986generalized};
however, this paper, along with several later works \cite{bender2018generalized, tsujitani2012survival, liu2018parametric, xu2023coxnam}, use the Cox loss and are thus constrained by the proportional hazards assumption. 
\cite{van2023interpretable} fits classification GAMs for survival analysis without assuming proportional hazards by using survival stacking \cite{craig2021survival}.
However, survival stacking is computationally challenging for larger datasets, and the method in \cite{van2023interpretable} requires feature selection as a distinct preprocessing step.
TimeNAM \cite{peroni2022extending} fits \GAM s with both proportional and non-proportional hazards losses, but uses a method similar to survival stacking that does not scale well to larger datasets.
PseudoNAM \cite{rahman2021pseudonam} fits GAMs using pseudo-values as labels for the survival distribution, which is similar to our method but requires an extra processing step to generate pseudo-values before training, which can be computationally expensive.

\section{Experiments}

Our empirical results demonstrate the utility of DyS as an interpretable survival analysis model. 
For all experiments, we run 10 trials, each trial with a different random seed.
We split our datasets with an 80/20 train/test split, and further designated 20\% of the training data as validation data.
When training deep learning models, we train models with the Adam optimizer and a learning rate of $10^{-4}$ for a total of 200 epochs, and stop trials if their performance on the validation set does not improve for 5 straight epochs.
Our main evaluation metric is time-dependent area under the ROC curve (henceforth referred to as AUC), as implemented in the scikit-survival package \cite{polsterl2020scikit}. 
Time-dependent AUC measures the discrimination ability of a model at a provided set of evaluation times, accounting for the probability of censoring.
Further, the mean AUC across all times is calculated as a summary metric for the model's overall discrimination ability.
This mean AUC serves as our primary evaluation metric.

We perform three sets of experiments.
First, we use synthetic data to demonstrate the effectiveness of our RPS loss versus the traditional Cox loss.
Second, we use a collection of (smaller) benchmark survival analysis datasets to demonstrate the effectiveness of DyS in terms of discrimination compared to existing state-of-the-art survival models.
Last, we use a large observational healthcare dataset  for heart failure prediction to demonstrate the scalability of DyS, as well as to evaluate DyS as a feature selection method.
We provide code to reproduce our results \footnote{\url{https://anonymous.4open.science/r/dys_paper_anon_code}}. 

% Datasets:
% \begin{itemize}
%     \item See Slack note.
% \end{itemize}

% Baselines:
% \begin{itemize}
%     \item Cox
%     \item RSF (when possible).
%     \item PseudoNAM (should beat because doesn't use interactions).
%     \item TimeNAM
%     \item CoxNAM
%     \item NAM (no interactions) on CoxPH loss.
%     \item DyS on CoxPH loss.
%     \item NAM (no interactions) on our loss.
%     \item DyS on our loss.
%     \item Maybe compare with and without feature selection? On the smaller datasets at least.
%     \item Maybe one other deep learning model?
%     \begin{itemize}
%         \item DeepHit?
%         \item DeepWeiSurv?
%         \item DeepSurv?
%         \item DPWTE?
%         \item Deep Recurrent Survival Analysis?
%     \end{itemize}
%     \item Multi-Task Logistic Regression?
%     \item Normal Grad-boosting Cox model?
%     \item 
% \end{itemize}

% Metrics:
% \begin{itemize}
%     \item C-index (Harrell)
%     \item Time-dependent AUC (sksurv)
%     \item Time-dependent Cox (Antolini)?
%     \item Brier score (sksurv)
% \end{itemize}

\subsection{Synthetic Data}
\label{sec:exp_synth}

\begin{figure*}
    \centering
    \begin{minipage}{.5\textwidth}
        \centering
        \includegraphics[width=\textwidth]{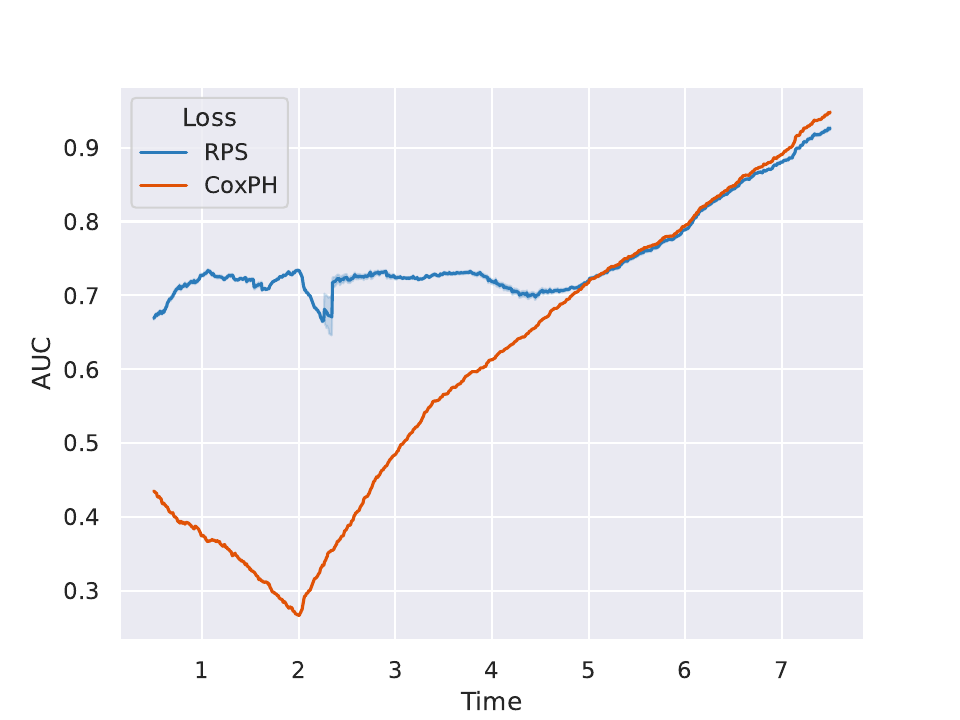}
    \end{minipage}%
    \begin{minipage}{.5\textwidth}
        \centering
        \includegraphics[width=\textwidth]{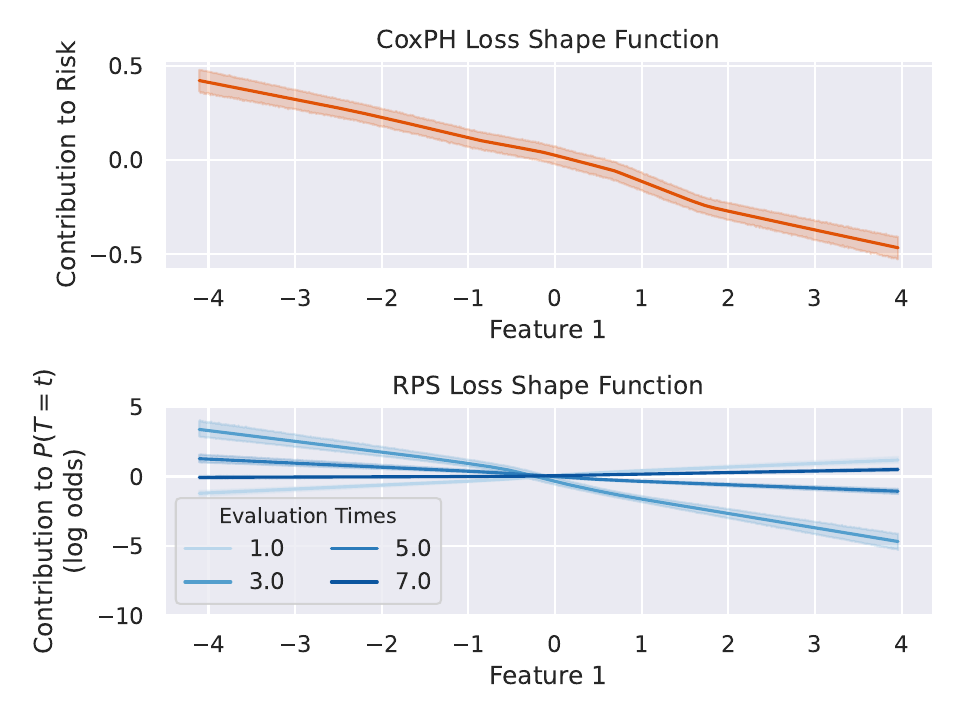}
    \end{minipage}
    \caption{Performance of DyS (with RPS loss) versus s CoxDyS, i.e. DyS using the CoxPH loss, on synthetic data which fails the proportional hazards assumption. (Left) Time-dependent AUC measured as several evaluation times, with dotted lines representing mean AUC. Using the CoxPH loss results in poor performance for smaller evaluation times. (Right) Shape functions for feature 1 under different loss functions. For RPS loss (bottom), shape function is shown at four different evaluation times, since DyS outputs are time-dependent.}
    \label{fig:synth}
\end{figure*}

We start with an experiment on synthetic data to demonstrate the utility of DyS as an accurate yet interpretable survival model. 
We generate synthetic data motivated by the example given in Section \ref{sec:loss_fn}; 
to recall, a child cancer patient may have higher mortality risk at very early or very late times, 
while an adult may have higher mortality risk at times in between.
This experiment shows that replacing the RPS loss in DyS with the Cox loss (which we will call CoxDyS, similar to CoxNAM \cite{xu2023coxnam}) yields inferior results.
We construct synthetic data such that each sample $X \in \R^p$ belongs to one of two groups with distinct time-to-event windows.
Group membership is decided by a linear function $Y = X^T \beta + \epsilon$, where smaller values of $Y$ correspond to group 1 and larger values of $Y$ correspond to group 2.
In group 1, the time-to-event variable $T \mid X \sim \mathcal{U}[2, 6]$, where $\mathcal{U}$ is the uniform distribution.
In group 2, $T \mid X \sim \mathcal{U}[0, 2]$ with probability 0.5, and $T \mid X \sim \mathcal{U}[6, 8]$ with probability 0.5. 
In words, samples in group 1 have moderate event times, while samples in group 2 have extreme event times, either late or early.
For complete details, see Appendix \ref{sec:synth_data}.
This synthetic data fails the proportional hazards assumption: 
$\lambda(t \mid X, \text{group} = 1) = 0$ for $t < 2$, 
whereas $\lambda(t \mid X, \text{group} = 2) = 0$ for $2 < t < 6$.
% Values of the feature representative of group 1 should result in higher risk at early and late times,
% and values representative of feature 2 should result in higher risk at intermediate times.

In Figure \ref{fig:synth}, we evaluate DyS versus CoxDyS on this synthetic data.
For each method, we plot the time-dependent AUC as a function of time to illustrate how each model performs at different times in the time-to-event horizon.
As expected, CoxDyS performs much worse, as the data violates proportional hazards.
Specifically, CoxDyS and DyS perform similarly for later times, but DyS performs much better than CoxDyS for earlier times.
This discrepancy is a direct result of the loss function: 
the proportional hazards loss ranks the samples in terms of risk independently of time, while the RPS loss used in DyS can change the sample rankings over time, which is necessary to correctly model the data.

Figure \ref{fig:synth} also shows shape functions generated by each approach for the first feature.
The coefficient for feature 1 in the data generating process, $\beta_1$, is 0.697, 
meaning smaller values of feature 1 are more likely to be in group 1, 
and larger values of feature 1 are more likely to be in group 2.
The shape functions for DyS reflect this behavior: 
at event times typical of group 1 ($t=3$ and $t=5$), small values of feature 1 contribute most to $P(T = t)$,
whereas at event times typical of group 2 ($t=1$ and $t=7$), large values of feature 1 contribute most.
Meanwhile, the single CoxDyS shape function does not accurately describe this behavior.
Instead, it displays a negative slope, which accurately reflects the risk only at some event times.

% The CoxDyS shape function plots the impact of feature 1 on time-independent risk, while the DyS shape function plots the impact of feature 1 on the survival probability at a given time $t$.
% This is superior to the CoxDyS shape functions, as the shape functions from DyS directly describe the impact to the quantity of interest (i.e. the survival curve), and can be time-specific.
% Generating time-specific shape functions is useful when the impact of a feature changes over time.
% This is true in our synthetic data, and is verified by the difference in shape functions for feature 1 generated by DyS at different times.

\subsection{Benchmark Datasets}
\label{sec:benchmark}

\begin{table*}
    \caption{Mean time-dependent AUC of DyS (with one-stage and two-stage fitting) compared to baselines on benchmark survival analysis datasets. All results are averaged over 10 trials, with standard deviations shown. Results in bold are within 1 standard deviation from the best result for each dataset. }
    \centering
    \begin{tabular}{lcccc}
    \toprule
    Dataset & flchain & metabric & mimic & support \\
    \midrule
    CoxPH & 0.948 ± 0.001 & 0.677 ± 0.011 & \textbf{0.674 ± 0.003} & 0.797 ± 0.001 \\
    RSF & \textbf{0.950 ± 0.001} & 0.732 ± 0.011 & 0.662 ± 0.004 & \textbf{0.819 ± 0.001} \\
    DeepSurv & \textbf{0.949 ± 0.001} & 0.698 ± 0.008 & \textbf{0.671 ± 0.003} & 0.798 ± 0.001 \\
    DeepHit & 0.944 ± 0.002 & 0.713 ± 0.011 & 0.656 ± 0.007 & 0.732 ± 0.016 \\
    DRSA & \textbf{0.946 ± 0.006} & 0.745 ± 0.007 & 0.661 ± 0.006 & \textbf{0.818 ± 0.002} \\
    SA Transformer & 0.945 ± 0.003 & 0.710 ± 0.012 & 0.659 ± 0.007 & 0.785 ± 0.006 \\
    PseudoNAM & 0.947 ± 0.001 & 0.641 ± 0.029 & 0.577 ± 0.012 & 0.814 ± 0.001 \\
    DyS (ours) one-stage & \textbf{0.951 ± 0.001} & \textbf{0.760 ± 0.004} & \textbf{0.674 ± 0.002} & \textbf{0.818 ± 0.001} \\
    DyS (ours) two stage & \textbf{0.949 ± 0.001} & \textbf{0.760 ± 0.003} & 0.669 ± 0.002 & 0.814 ± 0.001 \\
    \bottomrule
    \end{tabular}
    \label{tab:open_source_exp}
\end{table*}

We compare DyS to several existing survival models on a collection of standard survival analysis tasks.
We consider the following baseline models:
\begin{itemize}
    \item \textbf{CoxPH}: Standard linear Cox proportional hazards model \cite{cox1972regression}.
    \item \textbf{RSF}: Random Survival Forest \cite{ishwaran2008random}, an extension of random forests to survival analysis.
    \item \textbf{DeepSurv}: Deep learning model that uses the Cox loss \cite{katzman2018deepsurv}.
    \item \textbf{DeepHit}: Discrete-time deep learning model \cite{lee2018deephit}.
    \item \textbf{DRSA}: Deep Recurrent Survival Analysis, using a recurrent neural network for multi-timestep prediction \cite{ren2019deep}.
    \item \textbf{SA Transformer}: Transformer-based architecture for multi-timestep prediction \cite{hu2021transformer}.
    \item \textbf{PseudoNAM}: Interpretable neural additive model which uses pseudo-values as labels \cite{rahman2021pseudonam}.
\end{itemize}
We compare the above models to DyS on the following standard survival analysis datasets: flchain, metabric, mimic, support.
Dataset descriptions as well as preprocessing details can be found in Appendix \ref{sec:exp_details}.
Since these datasets all have a relatively small number of features (all have less than 100), we do not train DyS with any feature sparsity (i.e. we only use $L_{\text{RPS}}$ and not $L_{\text{SS}}$ or $L_{\text{E}}$).
Further, we train DyS with both one-stage and two-stage training to assess whether two-stage training is as effective as jointly training main effects and interactions.

The results are shown in Table \ref{tab:open_source_exp}. 
One-stage DyS is one of the leading performers on all 4 datasets, indicating that DyS is competitive with the state-of-the-art despite being interpretable.
Two-state DyS, as well, is either a leading performer or very closely behind on all datasets, providing evidence that two-stage fitting is very close to as good as fitting main effects and interactions jointly.
This is important because one-stage fitting is not possible on larger datasets, but we have evidence that two-stage fitting does almost as well when both are possible.
Additionally, PseudoNAM (the only other interpretable model) is competitive on flchain and support, but lags behind siginficantly on metabric and mimic.

\subsection{Heart Failure Prediction}
\label{sec:hf}

We evaluate DyS compared to baseline survival models for predicting heart failure risk.
We gather many features for a cohort of patients from a large hospital network (name censored for anonymity).
In total, our dataset has 2410 clinical features, including demographics, vital signs, lab results, conditions, and drug exposure, for a total of $\sim 670,000$ patients.  
Further details about cohort selection and preprocessing are in Appendix \ref{sec:hf_study_design}.

The scale of this dataset (and other similar observational healthcare datasets) presents distinct challenges.
First, datasets of this magnitude often necessitate feature selection (see Section \ref{sec:feature_selection}).
Second, models that do not scale well are simply too inefficient to be considered.
Specifically, in our experiments, the methods that do not use deep learning (CoxPH and RSF) are too slow (they do not finish in 10 hours). 
Last, interpretable models that consider all interaction terms are too slow; as discussed in Section \ref{sec:two-stage}, interaction screening like proposed in \cite{ibrahim2023grand} is too slow.

\subsubsection{Discrimination}

\begin{table}[]
    \centering
    \caption{Performance of DyS compared to baselines on the heart failure dataset. In the upper panel, all baselines use all available features (Cox is omitted due to inefficiency), while DyS does feature selection to select between 45 and 65 features. In the bottom panel, DyS and Cox use bisection (Algorithm \ref{alg:bisection} in Appendix \ref{sec:bisection}) to select exactly 10 features, while other baselines use the features generated by Cox.}
    \begin{tabular}{lcc}
    \toprule
    Dataset & Number of Features & Mean AUC \\
    \midrule
    Cox & all & --- \\
    DeepSurv & all & 0.536 ± 0.061 \\
    DeepHit & all & \textbf{0.829 ± 0.001} \\
    DRSA	& all	& 0.779 ± 0.008 \\
    SA Transformer	& all & 	0.812 ± 0.005 \\
    DyS (ours) & 45-65 & \textbf{0.826 ± 0.004} \\
    \midrule
    Cox & 10 & 0.769 ± 0.001  \\
    DeepSurv & 10 & 0.784 ± 0.000 \\
    DeepHit & 10 & 0.788 ± 0.000 \\
    DRSA &	10	& 0.762 ± 0.006 \\
    SA Transformer	& 10	& \textbf{0.798 ± 0.001} \\
    DyS (ours) & 10 & \textbf{0.799 ± 0.005} \\
    \bottomrule
    \end{tabular}
    \label{tab:hf_disc}
\end{table}

% \begin{table}[]
%     \centering
%     \caption{Caption}
%     \begin{tabular}{ll}
%     \toprule
%     dataset & heart failure \\
%     \midrule
%     DeepSurv & 0.536 ± 0.061 \\
%     NAM two stage & \textbf{0.826 ± 0.004} \\
%     DeepHit & \textbf{0.829 ± 0.001} \\
%     \bottomrule
%     \end{tabular}
%     \label{tab:hf_disc}
% \end{table}

We first compare DyS to baselines on the heart failure dataset in terms of discrimination ability.
As aforementioned, we omit CoxPH and RSF as they are too slow, and also omit PseudoNAM as computing pseudo-values is similarly slow.
For the remaining baselines (DeepSurv and DeepHit), we train similarly to Section \ref{sec:benchmark}.
For DyS, unlike Section \ref{sec:benchmark}, we use two-stage fitting (Algorithm \ref{alg:two-stage}) as well as feature selection via $L_{\text{SS}}$ and $L_{\text{E}}$ (Section \ref{sec:feature_sparsity}).
We set the regularization parameter $\lambda$ in $L_{\text{SS}}$ to a value that yields roughly 50 features, although we allow the number of features to vary across trials.

The results are shown in the top panel of Table \ref{tab:hf_disc}. DeepHit has the best average performance across trials, but DyS is within one standard deviation, despite using many less features than DeepHit.
This result demonstrates that the heart failure data can be approximated well by a small number of features, and DyS is successful at finding such a subset.

Further, interpretable plots for DyS models from this experiment are shown in Figure \ref{fig:interp_plots}.  
One interesting takeaway from these plots is that some features have feature impact functions with significantly different shapes at different evaluation times.
For example, the feature impact plot for body temperature has a positive slope at 1 year, but a negative slope at 3 years. 
One possible explanation for this behavior is that high body temperature corresponds to more immediate risk of heart failure, while low body temperature is correlated with some condition related to long-term heart failure risk.
The interaction plots also yield interesting findings; for one, patients with a short QRS duration (a feature related to electrocardiograms) as well as high systolic blood pressure seem to be at particularly high risk of heart failure in the near future.
Last, the feature importance plots confirm common understanding of the high correlation between age and heart failure risk, but also highlights the importance of gender and smoking, which are perhaps less well-known.

\subsubsection{Feature Selection}
\label{sec:select_k}

Next, we evaluate DyS as a feature selection method to select exactly $k$ features.
We set $k = 10$ as the number of features to select, and use bisection (Algorithm \ref{alg:bisection} in Appendix \ref{sec:bisection}) to find a hyperparameter setting that selects exactly 10 features.
For DyS, following Section \ref{sec:feature_selection}, we run bisection together with two-stage fitting, so that bisection is only done on the training of main effects.
The only baseline we compare to that is capable of feature selection is Cox, which can do feature selection via L1 regularization much like lasso regression \cite{simon2011regularization}.
Thus, for all other baselines, we first select a set of 10 features using bisection on an L1 penalized Cox model, and then train on the selected features.

The results are shown in the bottom panel of Table \ref{tab:hf_disc}.
While DeepHit and DyS had similar performance in the previous section, DyS now is clearly the superior model. 
This result emphasizes the importance of joint feature selection and predictive model, which DyS can do but DeepHit cannot.

% Things to include:
% \begin{itemize}
%     \item Table with discrimination metrics for DyS compared to baselines.
%     \begin{itemize}
%         \item All the baselines shouldn't need to select interactions, so shouldn't need to do feature selection.
%         \item For DyS, can optimize hyperparameters for sparsity on validation set, which might result in little sparsity.
%     \end{itemize}
%     \item Figures with shape functions and feature importances.
%     \item Table comparing using DyS vs using Cox + L1 as feature selection.
%     \begin{itemize}
%         \item Enforce that model can only using $k$ features (maybe $k = 10$), and run DyS with corresponding hyperparameters vs baselines run on 10 features selected from Cox + L1.
%         \item Rows in table:
%         \begin{itemize}
%             \item All baselines + Cox L1 features.
%             \item DyS using select-k regularization.
%             \item All baselines + DyS features? Maybe not necessary, start with other two...
%         \end{itemize}
%     \end{itemize}
% \end{itemize}

\section{Conclusion}

We present DyS, a new glass-box model for time-to-event data.
DyS is suitable for both small and large survival analysis datasets, and can be used both for interpretable prediction as well as for feature selection.
Our empirical results on benchmark survival analysis datasets demonstrate the utility of DyS as a general purpose survival analysis model.
Further, our results on a large heart failure survival task illustrate the  scalability and effectiveness of DyS for large-scale survival analysis problems, where other methods are either less effective or too slow.
We hope our work inspires further research into XAI, and particularly glass-box modeling, for survival analysis problems.

% \begin{acks}
% \end{acks}

\bibliographystyle{abbrv}
\bibliography{bib}

\appendix

\begin{table*}
    \centering
    \caption{Dataset descriptions of the datasets used in our experiments.}
    \begin{tabular}{lllll}
        \toprule
        Name & Domain & \# Samples & \# Features & \% censored \\
        \midrule
        flchain & Microbiology & 7874 & 44 & 72.4\% \\
        metabric & Genomics & 1981 & 79 & 55.2\% \\
        mimic & Healthcare & 15241 & 95 & 61.9\% \\
        support & Healthcare & 9105 & 57 & 31.9\% \\
        heart-failure & Healthcare & 672296 & 2410 & 97.5\% \\
        \bottomrule
    \end{tabular}
    \label{tab:datasets}
\end{table*}

\begin{table*}
    \centering
    \caption{Heart Failure Cohort Statistics.}
    \vspace{0.2cm}
    \scalebox{0.8}{
    \begin{tabular}{llll}
    \toprule
    \textbf{Total Features By Type}           & \textbf{Gender}         & \textbf{Race}             & \textbf{Age}                     \\
    \midrule
    Measurements: 1464 & Male: 39.5\%   & White: 54.2\%   &  18-29: 13.4\%           \\
    Conditions: 314    & Female: 60.5\% & Asian: 20.3\%   & 30-44: 25.9\%           \\
    Drugs: 614         &                & Black: 4.1\%    &  45-59: 27.5\%         \\
                       Demographics: 10         &                & Other: 1.2\%   & 60-74: 24.6\%         \\
    Observations: 8 &                & Unknown: 20.2\% & $\geq 75$: 8.6\% \\
    \bottomrule
    \end{tabular}
    }
    \label{tab:cohort_summary}
\end{table*}

\section{Additional Methodology Details}

\subsection{Smooth-Step Function}
\label{sec:smooth_step}

The smooth-step function $S(x; \mu)$ was originally proposed in \cite{hazimeh2020tree} as a neural network activation for training differentiable decision trees.
We follow \cite{ibrahim2023grand} as use it for learning sparse additive models.
The function is defined as
\begin{equation}
    S(x; \mu) = 
    \begin{cases}
        0 & \text{ if } t \leq -\mu/2 \\
        -\frac{2}{\mu^{3}}x^3 + \frac{3}{2\mu}t + \frac{1}{2} & \text{ if } -\mu/2 \leq x \leq \mu/2 \\
        1 & \text{ if } x \geq \mu/2
    \end{cases}
\end{equation}
The function can perform hard-selection by reaching exactly 0 or 1, while also being continuously differentiable.
The $\mu \in \R$ parameter in $S$ is the learnable parameter that the network can optimize to induce feature-sparsity during training.

\subsection{Bisection}
\label{sec:bisection}

Bisection is an algorithm traditionally used as an optimization search technique. 
In this paper, we use bisection when we want to find a hyperparameter, $\lambda$, for a feature selection procedure that results in exactly $k$ features. 
The algorithm is very similar to binary search, except that an initial upper and lower bound is not available. 
Thus, the algorithm starts by searching for hyperparameter settings $\lambda_{\text{low}}$ and $\lambda_{\text{high}}$ that result in too many and too few feature respectively, and then does a binary search until a feasible $\lambda$ is found. The algorithm is described fully in Algorithm \ref{alg:bisection}.

\begin{algorithm}
\caption{Select $k$ features by bisection.}
\begin{algorithmic}[1]
    \State Input
    \begin{itemize}[noitemsep, topsep=0pt]
        \item Feature selection model $f$, train dataset $D$.
        \item Initial regularization parameter $\lambda_0$.
        \item Target feature number $k$.
    \end{itemize}

    \State $\lambda_{\text{high}} \gets \text{None}$, $\lambda_{\text{low}} \gets \text{None}$.
    \State $\lambda \gets \lambda_0$.
    \State $\text{num}\_{\text{feats}} \gets -1$.

    \While{num feats $ \neq k$}
    
        \State $\text{num}\_{\text{feats}} \gets f(D)$.

        \If{num feats $< k$}
            \State $\lambda_{\text{high}} \gets \lambda$.
            \If{$\lambda_{\text{low}}$ is None}
                \State $\lambda \gets \lambda / 2$.
            \Else
                \State $\lambda \gets (\lambda_{\text{low}} + \lambda_{\text{high}}) / 2$.
            \EndIf

        \ElsIf{num\_feats $ > k$}
            \State $\lambda_{\text{low}} \gets \lambda$.
            \If{$\lambda_{\text{high}}$ is None}
                \State $\lambda \gets 2 \cdot \lambda$.
            \Else
                \State $\lambda \gets (\lambda_{\text{low}} + \lambda_{\text{high}}) / 2$.
            \EndIf

        \Else
            \State \Return $f$.
        \EndIf
    \EndWhile
    
\end{algorithmic}
\label{alg:bisection}
\end{algorithm}

\section{Additional Experiment Details}
\label{sec:exp_details}

Details about our datasets used can be found in Table \ref{tab:datasets}. 
We provide further details about each dataset below:
\begin{itemize}
    \item \textbf{flchain}: we obtain via scikit-survival \cite{polsterl2020scikit}.
    \item \textbf{metabric}: we obtain via the GitHub repository for DeepHit \footnote{https://github.com/chl8856/DeepHit/tree/master/sample\%20data/METABRIC}. We thus use the same preprocessing as from the DeepHit paper \cite{lee2018deephit}.
    \item \textbf{mimic}: we gather features from the MIMIC III dataset using the MIMIC benchmark \cite{Harutyunyan2019}. We follow the steps for generating features for the mortality logistic regression task, keeping only the features which consider the entire historical window. Then, we get the survival label for each patient following \cite{avati2020countdown}, and join on patient ID. Further details can be found in our code.
    \item \textbf{support}: we obtain following the DeepSurv paper \cite{katzman2018deepsurv}.
    \item \textbf{heart-failure}: see Section \ref{sec:hf_study_design}.
\end{itemize}

We use one-hot encoding for all categorical features, and standard scaling to zero mean and unit variance for all continuous features.
We replace missing values with the new category ``Unknown'' for categorical features, and with 0 (after scaling) for continuous features.

\subsection{Heart Failure Study Design}
\label{sec:hf_study_design}

We obtain data from the electronic health records (EHR) from a large hospital network (name censored for anonymity). The EHR data is structured in the OMOP Common Data Model (CDM) \cite{hripcsak2015observational}.

\subsubsection{Features}

We gather features from the following OMOP CDM tables:
\begin{itemize}
    \item \textbf{Person}: source for age, race, and gender.
    \item \textbf{Measurements}: source for vital sign and lab measurement features.
    \item \textbf{Drug Exposure}: source for exposure to different medications.
    \item \textbf{Condition Occurrence}: source for whether or not patients have certain medical conditions.
    \item \textbf{Observation Occurrence}: source for smoking data.
\end{itemize}
For each patient, we gather all values available for each feature across the patient's historical data, giving us a time-series for each feature.
For each continuous feature in the measurements table, we generate 6 tabular features from the feature's time series: the most recent value, as well as the mean, minimum, maximum, standard deviation, and count of non-missing values over the entire time series.
For each categorical feature from the measurements table, as well as for smoking, we generate 1 tabular feature using the last available measurement.
For each condition or drug exposure feature, we generate 2 binary features: 1) whether (1) or not (0) the patient has a record of the condition or drug in the last year (short term), or any time in their history (long term).

\subsubsection{Index Date}
For each patient, we designate their prediction date, or index date, as their visit interaction with the EHR after January 1, 2015. 
Further, we restrict our cohort to patients older than 18 with at least one year of observation before and after their index date, and who have not gotten heart failure before their index date.
For the survival label, we access whether or not a patient gets heart failure after their index date using a defined collection of concepts indicative of heart failure diagnosis.
All patients who do get heart failure are left uncensored, and their time-to-event label is the number of days between index date and heart failure diagnosis.

\subsection{Synthetic Data}
\label{sec:synth_data}

We generate synthetic data for the experiment in Figure \ref{fig:synth} using the following procedure. 
We start by generating features $X \in \R^p$, coefficients $\beta \in \R^p$, and errors $\epsilon \in \R^p$ from a standard normal distribution (we use $p = 10$ features).
Next, we calculate a continuous response $Y = X^T \beta + \epsilon$. 
We pass this response, which is normally distributed, through the CDF of a normal distribution with mean and variance given by the empirical mean and variance of $Y$.
The result of this transform, call it $Y^*$, is uniformly distributed between 0 and 1.
Then, we designate $Y^*$ into one of two groups, cutting at the median of the distribution of $Y^*$.
In group 1, we min-max scale $Y^*$ into range $[2T / 8, 6T / 8]$, where $T$ is the inputted maximum event time.
In group 2, we min-max scale $Y^*$ into range $[0, 2T / 8]$, and add $6 S / 8$ where $S$ is a binary random variable $S \sim \text{Bernoulli}(0.5)$.
The resulting random variable is approximately distributed uniformly in $[2T / 8, 6T / 8]$ with probability 0.5, and uniformly in $[0, 2T / 8] \cup [6T / 8, 1]$ with probability 0.5, as described in Section \ref{sec:exp_synth}.

\subsection{Model Implementations}

\begin{itemize}
    \item \textbf{CoxPH}: we fit CoxPH using the implementation in scikit-survival. 
    We use L2 regularization with regularization parameter of $10^{-3}$ for all experiments to improve numerical stability.
    \item \textbf{RSF}: we use the implementation in scikit-survival, with default hyperparameters.
    \item \textbf{DeepSurv}: we base our implementation off of \url{https://github.com/czifan/DeepSurv.pytorch}, a PyTorch implementation of the original DeepSurv repository which is writte in Tensorflow.
    We use a hidden dimension of 128 and no dropout or L2 regularization for all experiments except heart failure, for which we use a dropout of 0.2 and L2 regularization of $10^{-3}$ to combat overfitting.
    \item \textbf{DeepHit}: we use the implementation in PyCox \url{https://github.com/havakv/pycox}, a package with several deep learning survival models in PyTorch.
    We use a hidden dimension of 128, and dropout of 0.6, following the original paper.
    Last, we use 100 evaluation output times.
    \item \textbf{DRSA}: we write our own implementation based on the DRSA paper \cite{ren2019deep}. We use a hidden dimension of 128 and 100 evaluation times.
    \item \textbf{SA Transformer}: we write our own implementation based on the Github code accompanying the original paper \cite{hu2021transformer}. We use the hyperparameters suggested in the Github.
    \item \textbf{PseudoNAM}: we generate pseudo-values using the Kaplan-Meier estimator from scikit-survival.
    Each shape function is implemented as an MLP with hidden dimension 32. 
    We use 100 evaluation times, i.e. 100 pseudo-value labels.
    \item \textbf{DyS}: we use a hidden dimension of 32 for all shape function MLPs, and 100 evaluation times.
\end{itemize}

\end{document}